\documentclass{article}

\usepackage[final]{neurips_2024}

\makeatletter
\renewcommand{\@noticestring}{Proceedings of the LatinX in AI Workshop @ NeurIPS-25.}
\makeatother

\usepackage{tabularx}
\usepackage{natbib}

\usepackage[utf8]{inputenc} 
\usepackage[T1]{fontenc}    
\usepackage{hyperref}       
\usepackage{url}            
\usepackage{booktabs}       
\usepackage{amsfonts}       
\usepackage{nicefrac}       
\usepackage{microtype}      
\usepackage{xcolor}         
\usepackage{graphicx}
\usepackage{subcaption}
\usepackage{float}        
\usepackage{placeins}     

\title{Explaining AI Without Code: A User Study on Explainable AI}

\author{
  Natalia Abarca \\
   Department of Computer Science, University of Chile \\
  \texttt{natalia.abarca@ug.uchile.cl}
  \And
  Andrés Carvallo \\
  CENIA – National Center for Artificial Intelligence \\
  \texttt{andres.carvallo@cenia.cl}
  \And
  Claudia López Moncada \\
  Universidad Técnica Federico Santa María \\
  CENIA – National Center for Artificial Intelligence\\
  \texttt{clopez@inf.utfsm.cl}
  \And
  Felipe Bravo-Marquez \\
  Department of Computer Science, University of Chile \\
  CENIA – National Center for Artificial Intelligence\\
  IMFD - Millennium Institute for Foundational Research on Data\\ 
  \texttt{fbravo@dcc.uchile.cl}
}

\begin{document}

\maketitle

\begin{abstract}
The increasing use of Machine Learning (ML) in sensitive domains such as healthcare, finance, and public policy has raised concerns about the transparency of automated decisions. Explainable AI (XAI) addresses this by clarifying how models generate predictions, yet most methods demand technical expertise, limiting their value for novices. 
This gap is especially critical in no-code ML platforms, which seek to democratize AI but rarely include explainability. We present a human-centered XAI module in \href{https://www.dash-ai.com/}{DashAI}, an open-source no-code ML platform. The module integrates three complementary techniques, which are Partial Dependence Plots (PDP), Permutation Feature Importance (PFI), and KernelSHAP, into DashAI’s workflow for tabular classification. 
A user study ($N=20$; ML novices and experts) evaluated usability and the impact of explanations. Results show: (i) high task success ($\geq80\%$) across all explainability tasks; (ii) novices rated explanations as useful, accurate, and trustworthy on the Explanation Satisfaction Scale (ESS, Cronbach’s $\alpha$ = 0.74, a measure of internal consistency), while experts were more critical of sufficiency and completeness; and (iii) explanations improved perceived predictability and confidence on Trust in Automation (TiA, $\alpha$ = 0.60), with novices showing higher trust than experts. These findings highlight a central challenge for XAI in no-code ML, making explanations both accessible to novices and sufficiently detailed for experts.
\end{abstract}

\section{Introduction}

Artificial Intelligence (AI) systems are increasingly integrated into everyday life, assisting decision-making in various domains, including healthcare \citep{Alowais2023Revolutionizing}, finance \citep{cao2022ai}, law \citep{magesh2025hallucination}, and personal assistants \citep{rillig2024ai}. 

However, the widespread use of complex \textit{``black-box''} models, whose internal mechanisms are not interpretable to humans, raises critical concerns about transparency \citep{Thalpage2023Unlocking, Franzoni2023From, Hassija2023Interpreting}, accountability \citep{Ananny2018Seeing,
Burkart2020A}, and trust \citep{Wanner2022The}, particularly in sensitive contexts where errors or biases can have severe consequences. This issue has increased the growth of Explainable Artificial Intelligence (XAI), a field focused on making machine learning (ML) models more interpretable and their predictions understandable \citep{Naser2021An}. 

XAI aims both to help humans comprehend and trust AI outcomes and to support model development by exposing potential biases or errors \citep{ribeiro2016should,lundberg2017unified}. Despite advances, most methods remain isolated tools that require programming expertise, limiting adoption in real-world workflows \citep{Karim2022Explainable}. 

In parallel, \textit{no-code ML solutions} have emerged to democratize AI by allowing users to train and deploy models through graphical interfaces \citep{Sundberg2023Democratizing,
Li2022How}. Such tools broaden access to ML, enabling a heterogeneous user base that includes novices, domain experts, and ML practitioners. Nevertheless, most of these systems lack built-in explainability, creating an \textit{explainability gap}: they lower barriers to model creation but leave users without means to understand, validate, or trust predictions \citep{Burkart2020A,
Bhatt2019Explainable, 
Belle2020Principles}. The problem is exacerbated by user diversity, as novices seek transparency and trust while experts demand deeper inspection and diagnostic capabilities \citep{Herm2022Stop,
Marcinkevics2023Interpretable}. Recent research emphasizes human-centered approaches to XAI, ensuring explanations are not only technically correct but also comprehensible and helpful across user groups \citep{Kong2024Toward,
Al-Ansari2024User‐Centered,
Liao2021Human-Centered}. Embedding multiple complementary methods directly into the ML workflow can bridge the gap between theoretical advances in XAI and practical adoption \citep{Guo2024Explainability}. 

In this paper, we present the design, implementation, and evaluation of an interactive explainability module integrated into \href{https://www.dash-ai.com/}{DashAI}, a no-code ML software. The system incorporates three XAI techniques: Partial Dependence Plots (PDP) \citep{Hooker2019Unrestricted}, Permutation Feature Importance (PFI) \citep{Molnar2020Model-agnostic}, and KernelSHAP \citep{Aas2019Explaining}, to provide global and local explanations of tabular classification models. 

We validate the module through a user study ($N=20$) with novices and ML experts, evaluating usability, satisfaction, and trust in the explanations. 

\noindent The contributions of this paper are as follows:
\begin{itemize}
    \item Integration of three complementary XAI methods (PDP, PFI, KernelSHAP) into a no-code ML workflow. 
    \item A user study with $N=20$ indicating a task success ratio, a System Usability Score (SUS) and overall positive satisfaction across groups.
    \item Insights into differences between novices and experts, with implications for future human-centered XAI design. 
\end{itemize}

The remainder of this paper is organized as follows: Section~\ref{relatedwork} reviews related work in explainable AI toolkits and interfaces. Section~\ref{system} presents the system and its explainability module. Section~\ref{userstudy} details the user study design. Section~\ref{results} reports the results, and Section~\ref{conclusions} makes conclusions with future directions.  

\section{Related Work}
\label{relatedwork}

Research in Explainable Artificial Intelligence (XAI) has yielded a diverse range of methods, libraries, and interfaces designed to enhance the transparency of machine learning models. However, the extent to which these approaches integrate into no-code environments and how different user groups perceive them remains an open question.

Building on this foundation, several empirical studies have investigated how humans interact with explanations of machine learning systems \citep{Narayanan2018How}. Early work has shown that local explanation methods such as LIME \citep{ribeiro2016should} and SHAP \citep{sundararajan2020many} can improve users’ ability to understand model predictions \citep{Sathyan2022Interpretable, 2023Interpretable, Chiesa-Estomba2023Explainable}. Subsequent research examined how explanation quality impacts user trust, satisfaction, and reliance on automated decisions, but results remain mixed \citep{Papenmeier2022It’s, Pareek2024Effect, Westphal2023Decision}. More recent studies have highlighted the role of visualization in shaping interpretability and trust, particularly through attention-based methods in text classification and medical contexts \citep{parra2019analyzing, carvallo2025user_final}. Meanwhile, tools like Tsundoku demonstrate how explainability can be embedded into human-centered AI systems \citep{graells2025tsundoku}.

Moving beyond technical correctness, scholars have increasingly emphasized the usability of XAI as a key concern. Human–Computer Interaction (HCI) research emphasizes that explanations must not only be accurate but also understandable and useful for decision-making \citep{Picard2023Human-Computer, Reddy2024Human-Computer}. To this end, several studies have proposed design guidelines and evaluation frameworks for human-centered XAI, incorporating measures such as the System Usability Scale (SUS), Explanation Satisfaction Scale (ESS), and Trust in Automation (TiA) \citep{Kong2024Toward, Mohseni2018A, Rong2022Towards}. These metrics enable systematic evaluation of how explanations influence user confidence, understanding, and reliance on AI systems.

In parallel with these efforts, no-code machine learning platforms have sought to democratize AI by allowing users to train and deploy models without programming. Platforms such as Google Cloud AutoML, H2O Driverless AI, Orange3, KNIME, and RapidMiner are increasingly used by business analysts, domain experts, and educators. However, explainability in these systems is often restricted to simple global feature importance or static plots, without interactive or user-centered explanation modules \citep{Tian2024Automated}. Only a small number of studies have explicitly examined XAI in no-code or low-code settings, limited to a survey on natural language explanations \citep{Cambria2023A} and the XEdgeAI framework for human-centered industrial inspection \citep{Nguyen2024XEdgeAI}. While there is a clear and growing need for robust explanation capabilities in these no-code platforms, research efforts to address this demand remain scarce. Additionally, there is a lack of empirical evidence on how diverse user groups actually interact with explainability modules in no-code environments. 

Most prior work has concentrated on isolated evaluations of single explanation methods \citep{Bodria2021Benchmarking, Waa2021Evaluating}, and rigorous comparisons of novices and experts remain scarce. Addressing this gap is crucial to ensure that explainability in no-code ML platforms is not only technically accurate but also usable, trustworthy, and accessible across heterogeneous audiences.

\section{System Description}
\label{system}

\subsection{Overview of DashAI}

\begin{figure}[!htbp]
    \centering
    \includegraphics[width=0.7\textwidth]{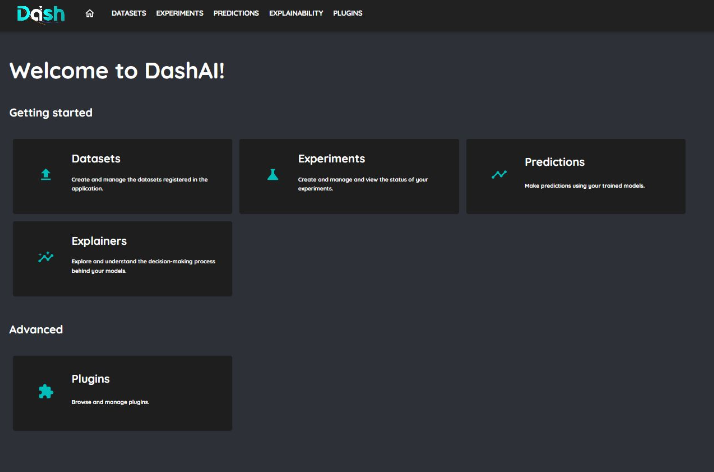}
    \caption{Home view of DashAI. Users can access datasets, create experiments, run predictions, explore explainability tools, and manage plugins.}
    \label{fig:dashai-home}
\end{figure}

\href{https://www.dash-ai.com/}{DashAI} is an open-source no-code machine learning (ML) software to democratize access to AI technologies for a diverse range of users. The system provides a graphical user interface that guides users through the complete ML workflow without requiring programming skills or advanced expertise in statistics or computer science. 
The home screen (Figure~\ref{fig:dashai-home}) offers access to the core modules: \textit{Datasets}, \textit{Experiments}, \textit{Predictions}, \textit{Explainers}, and \textit{Plugins}. This design emphasizes usability, allowing novices and domain experts to orient themselves quickly and navigate the available functionalities.

\begin{figure}[ht]
    \centering
    \begin{subfigure}[b]{0.6\textwidth}
        \centering
        \includegraphics[width=\linewidth]{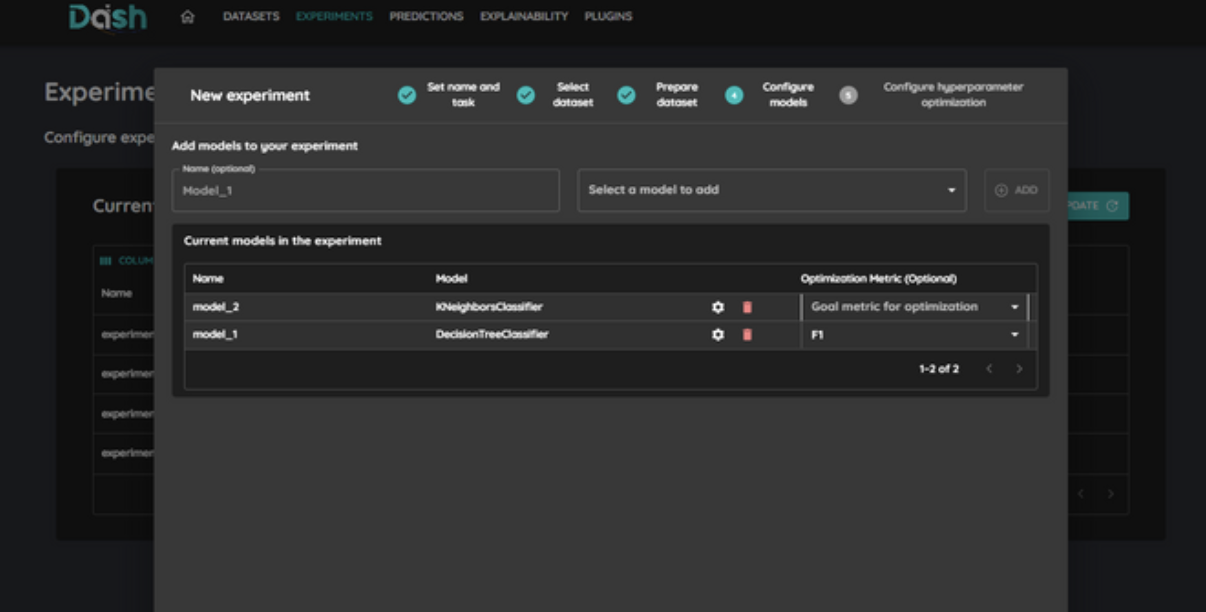}
        \caption{Experiment configuration and training module.}
        \label{fig:exp-module}
    \end{subfigure}
    
    \begin{subfigure}[b]{0.6\textwidth}
        \centering
        \includegraphics[width=\linewidth]{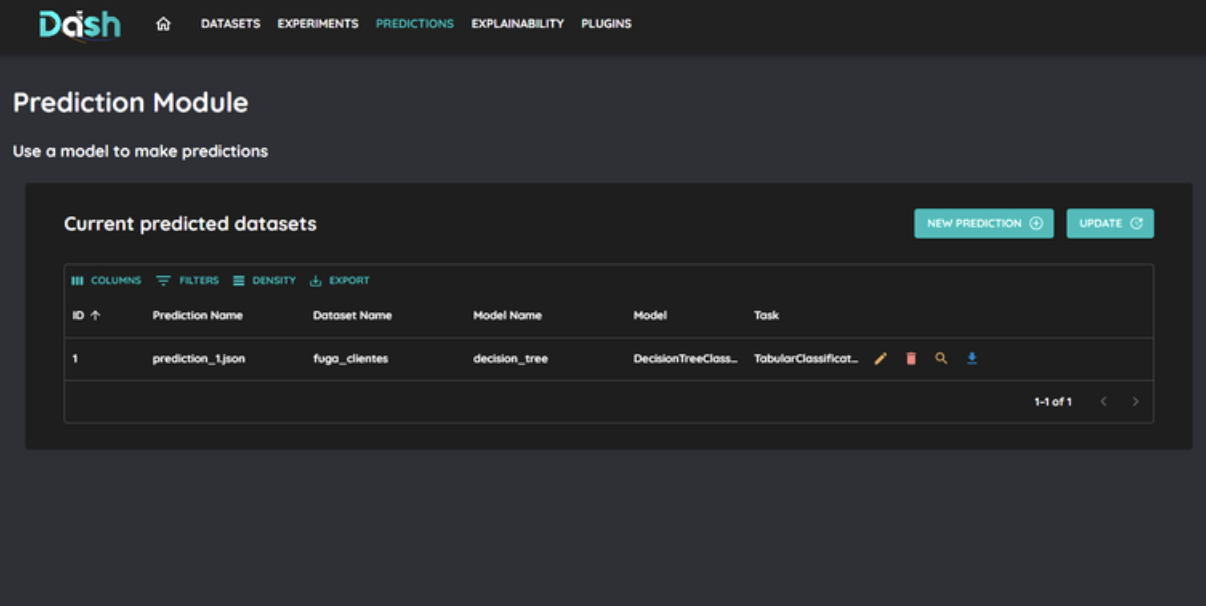}
        \caption{Prediction module for applying trained models to new data.}
        \label{fig:pred-module}
    \end{subfigure}
    \caption{Experiment and prediction modules in DashAI. Users configure and train models (a), and subsequently apply them to unseen data for prediction (b).}
    \label{fig:dashai-exp-pred}
\end{figure}

Once a dataset is uploaded, users can configure and train models through the \textit{Experiments} module. Figure~\ref{fig:dashai-exp-pred}(a) shows the experiment setup interface, where users select algorithms, adjust configurations, and track models within an experiment. Trained models can then be deployed to generate predictions, as illustrated in Figure~\ref{fig:dashai-exp-pred}(b), which presents the \textit{Predictions} module summarizing results for new datasets.

\subsection{Explainability Module}

The explainability module was designed to provide both global and local insights into trained models. It integrates three complementary techniques: Partial Dependence Plots (PDP), Permutation Feature Importance (PFI), and KernelSHAP. These are embedded directly into the workflow so that users can interpret their models immediately after training, without relying on external tools.

\begin{figure}[!t]
    \centering

    \begin{subfigure}[b]{0.5\textwidth}
        \centering
        \includegraphics[width=0.95\linewidth]{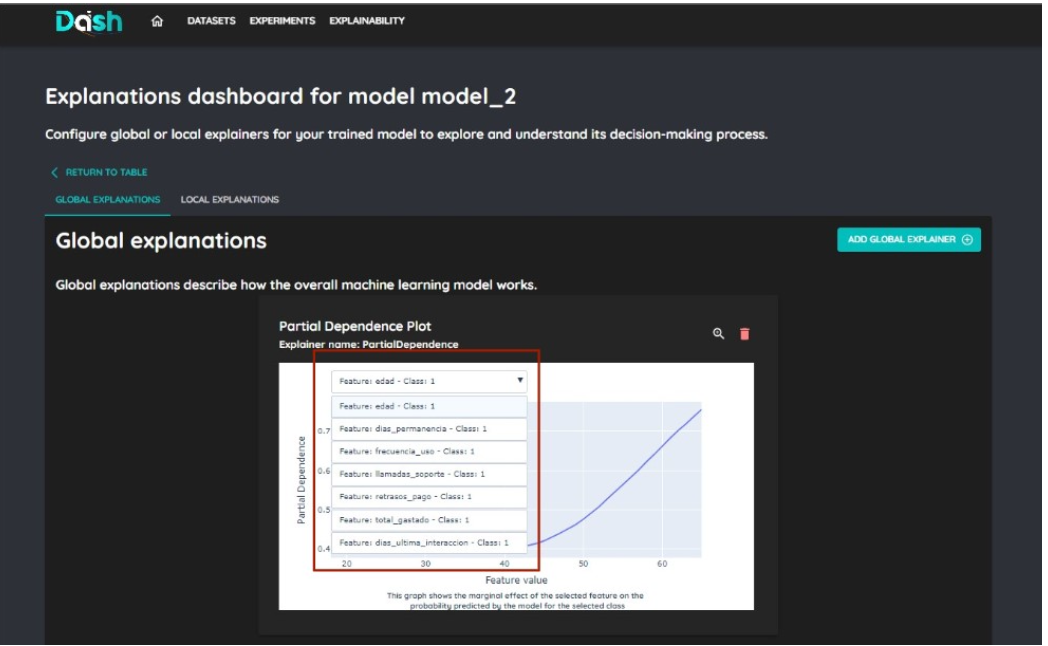}
        \caption{Partial Dependence Plot (PDP).}
        \label{fig:pdp}
    \end{subfigure}
    \hfill
    \begin{subfigure}[b]{0.45\textwidth}
        \centering
        \includegraphics[width=0.95\linewidth]{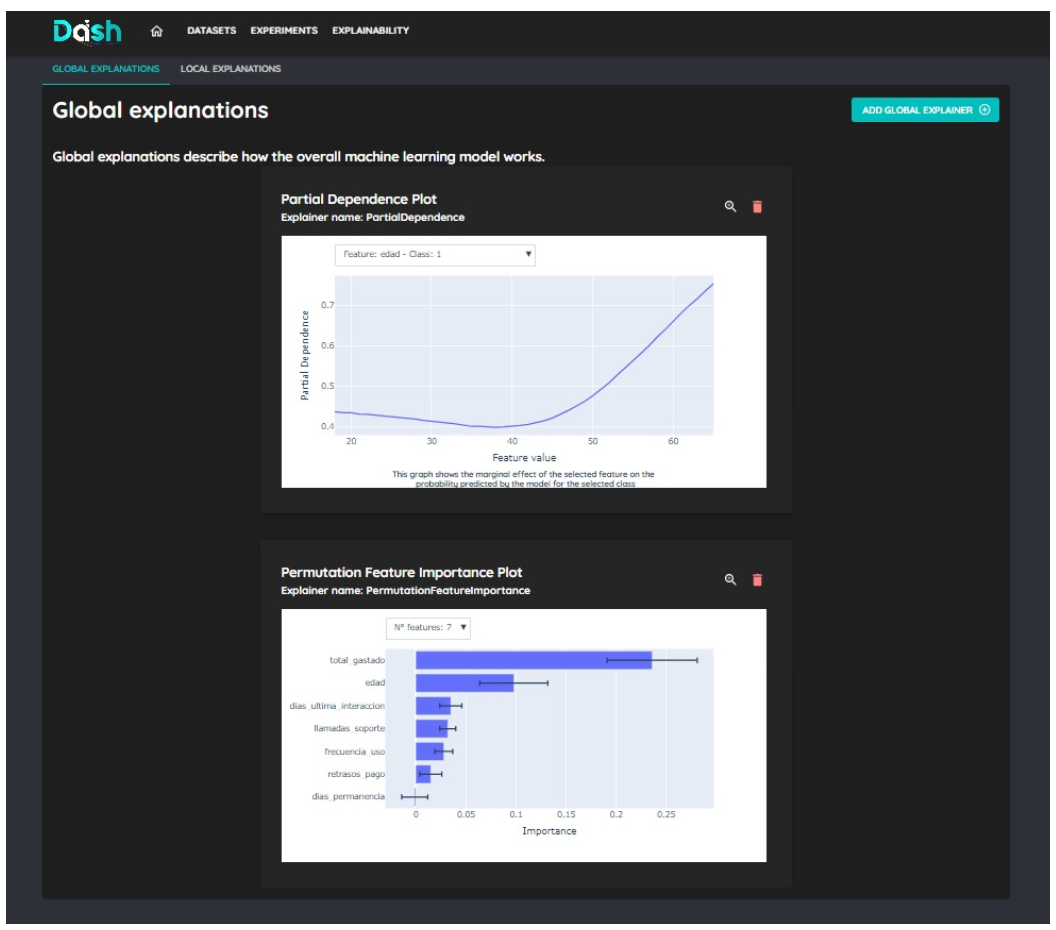}
        \caption{Permutation Feature Importance (PFI).}
        \label{fig:pfi}
    \end{subfigure}

    \vspace{0.5cm}

    \begin{subfigure}[b]{0.6\textwidth}
        \centering
        \includegraphics[width=\linewidth]{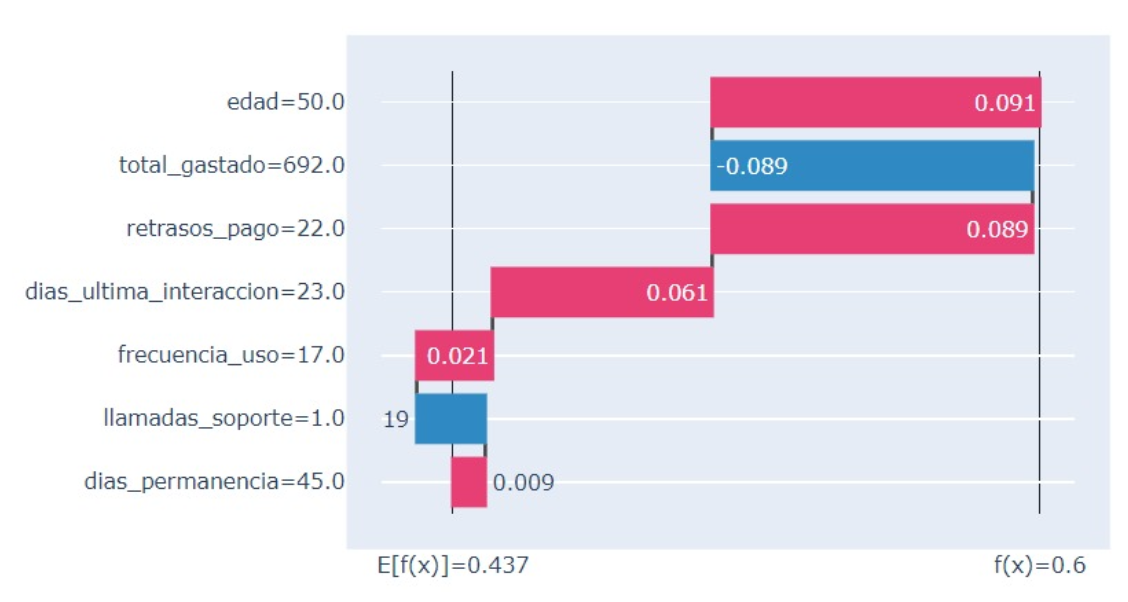}
        \caption{SHAP local explanation.}
        \label{fig:shap}
    \end{subfigure}
    
    \caption{Explainability module in DashAI. PDP (a) and PFI (b) provide global insights into model behavior, while SHAP (c) illustrates local instance-level explanations.}
    \label{fig:dashai-xai}
\end{figure}

These modules are the focus of the user study in the next section. 
Figure~\ref{fig:dashai-xai} summarizes the explanation methods integrated into DashAI. 
\textit{PDP} (Fig.~\ref{fig:dashai-xai}a) shows how changes in a single feature affect the predicted probability of the target class. 
\textit{PFI} (Fig.~\ref{fig:dashai-xai}b) ranks features by estimating the performance drop when each variable is randomly permuted. 
Together, these global explanations provide an overview of model behavior and highlight the most influential variables. 
In addition, \textit{KernelSHAP} (Fig.~\ref{fig:dashai-xai}c) offers local explanations by decomposing an individual prediction into feature-level contributions, 
helping users understand why the model produced a specific outcome for a given instance.

\section{User Study}
\label{userstudy}

\subsection{Study Design}
We conducted a between-subjects user study with $N=20$ participants (Novices $n=10$, Experts $n=10$) to evaluate the \textit{explainability module} of DashAI. Participants were randomly assigned to two experimental conditions: 
\textit{Scenario A} (progressive analysis, where explanations were generated and analyzed sequentially: PDP, PFI, and KernelSHAP) and \textit{Scenario B} (combined analysis, where the three explanations were generated and analyzed together). Each scenario included 5 novices and 5 experts. 

The overall user test originally consisted of eleven tasks (see Annex \ref{annex:tasks}). Tasks T1–T6 covered general workflow steps such as loading datasets, training models, and exploring results, which are necessary prerequisites for obtaining explanations. However, for the purposes of this work we focus exclusively on Tasks T7–T11, which directly involve the explainability module. 

The study goals were twofold: (i) assess the usability of the explainability interface, and (ii) measure user satisfaction with the explanations. Each session lasted approximately one hour and included task execution, administration of questionnaires and a brief semi-structured interview (see Annex \ref{annex:structured_interview}). 

Participants completed the following explanation-related tasks:
\begin{enumerate}
    \item Open the explainability dashboard.
    \item Generate and interpret a PFI explanation.
    \item Generate and interpret a PDP explanation.
    \item Generate and interpret a KernelSHAP explanation for a specific instance.
    \item Generate and compare the three methods together.
\end{enumerate}

\subsection{Evaluation Metrics}
\begin{itemize}
    \item \textbf{Task success:} measured as the percentage of participants completing each of the five explainability tasks successfully (Annex \ref{annex:tasks}).
    \item \textbf{Satisfaction with explanations:} measured using the Explanation Satisfaction Scale (ESS) \citep{hoffman2018metrics}, which evaluates eight dimensions of explanation quality: comprehension, satisfaction, sufficiency of details, completeness, usability, usefulness, accuracy, and trust (see Annex \ref{annex:ess}).
    \item \textbf{Trust in automation:} measured using four subscales of the Trust in Automation (TiA) questionnaire (Annex \ref{annex:tia}), covering familiarity, predictability, propensity to trust, and confidence in automation. This scale complements ESS by focusing specifically on how explainability influences user confidence in AI-assisted decision-making.
\end{itemize}

\section{Results}
\label{results}

\paragraph{Task success.}
All explainability tasks achieved $\geq$80\% success (Table~\ref{tab:tasks-success}), indicating good usability of the module. Most errors occurred in T9–T11, typically during generation, analysis and comparison for local (KernelSHAP) and global explanations.

\begin{table}[ht]
\centering
\caption{Explainability tasks: success rates ($N=20$).}
\label{tab:tasks-success}
\begin{tabular}{llc}
\toprule
\textbf{Task ID} & \textbf{Description} & \textbf{Success (\%)} \\
\midrule
T7 & Open explainability dashboard & 100 \\
T8 & Generate and analyze PFI & 100 \\
T9 & Generate and analyze PDP & 90 \\
T10 & Generate and analyze KernelSHAP & 80 \\
T11 & Compare PDP, PFI, and KernelSHAP & 90 \\
\bottomrule
\end{tabular}
\end{table}

\paragraph{Reliability and statistical methods.}
To ensure validity of questionnaire data, we first assessed internal consistency of the ESS and TiA using Cronbach’s alpha, a standard reliability coefficient. Values above $\alpha=0.7$ are commonly considered acceptable, while lower values suggest caution in interpretation. In addition, group comparisons were conducted with non-parametric Mann–Whitney U tests (suitable for small, independent groups), and novice–expert differences were further explored through logistic regression with a median-split transformation of average scores.

\paragraph{Satisfaction with explanations (ESS).}
The ESS achieved $\alpha=0.74$, indicating acceptable reliability. Median scores showed strong agreement with usefulness and accuracy ($\tilde{X}=5$), while comprehension, satisfaction, sufficiency of details, completeness, usability, and trust obtained a median of 4. This suggests that participants found explanations helpful, reasonably detailed, and trustworthy.

A Mann–Whitney U test showed no significant differences between progressive (Scenario A) and combined (Scenario B) workflows ($U=38$, $p=0.38$). Logistic regression comparing novices and experts yielded a marginally significant effect ($p=0.082$), with novices rating explanations more positively overall. Novices emphasized transparency and reassurance, whereas experts were more critical of sufficiency of details and completeness.

\begin{figure}[ht]
    \centering
    \begin{minipage}{0.48\textwidth}
        \centering
        \includegraphics[width=\linewidth]{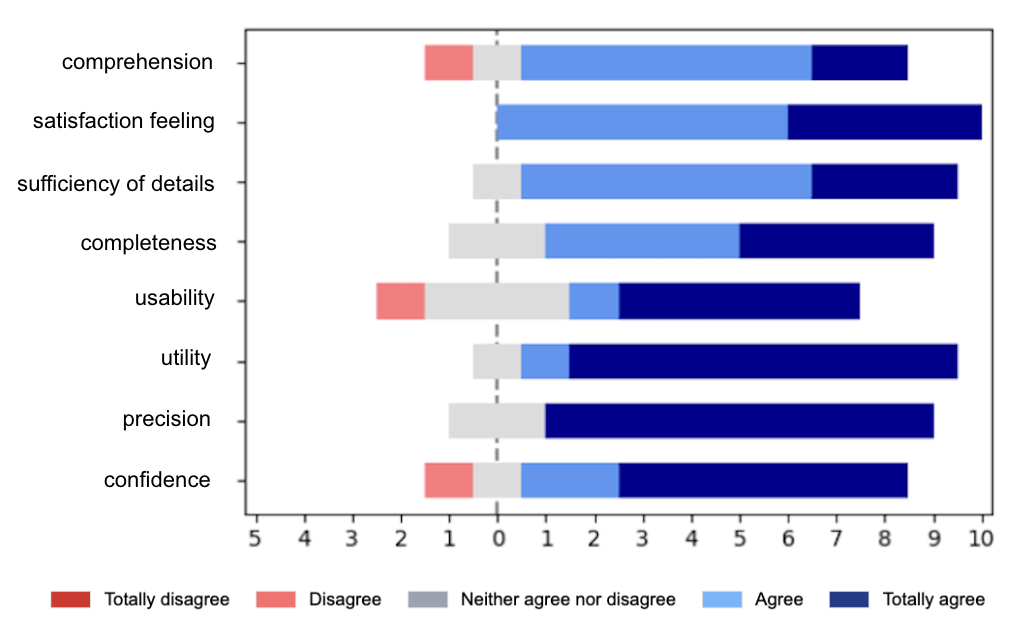}
        \caption*{(a) Novices ($N=10$)}
        \label{fig:ess-novices}
    \end{minipage}\hfill
    \begin{minipage}{0.48\textwidth}
        \centering
        \includegraphics[width=\linewidth]{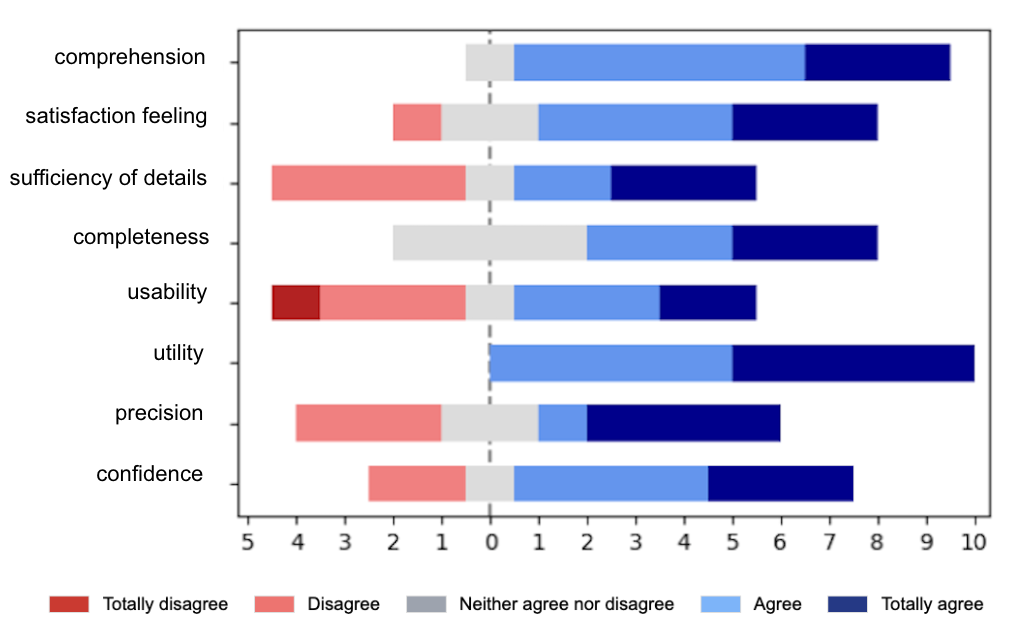}
        \caption*{(b) Experts ($N=10$)}
        \label{fig:ess-experts}
    \end{minipage}
    \caption{ESS responses by user group. Diverging Likert with disagreement (left) and agreement (right).}
    \label{fig:ess-combined}
\end{figure}

Figures~\ref{fig:ess-combined} illustrate these results. Novices (Fig.~\ref{fig:ess-combined}a) concentrated their responses in \textit{Agree/Strongly Agree}, particularly on usefulness, accuracy, and trust. Experts (Fig.~\ref{fig:ess-combined}b) displayed greater variability and some disagreement, consistent with their higher expectations for diagnostic depth.

\paragraph{Trust in automation (TiA).}
The TiA subscales yielded $\alpha=0.60$, indicating questionable reliability. While results should be interpreted with caution, they provide complementary insights. Participants reported moderate to high agreement on predictability and confidence, suggesting that explanations improved their ability to judge when to rely on the model. Familiarity and propensity to trust varied more strongly: novices tended to trust the system more readily, while experts were more reserved. These exploratory findings indicate that explainability not only supports understanding but also shapes confidence in AI-assisted decision-making.

In sum, the study shows that:  
(i) the explainability module was usable, with high task success rates ($\geq$80\%);  
(ii) ESS results confirmed high perceived usefulness, accuracy, and moderate trust, with novices rating explanations more positively than experts; and  
(iii) TiA responses, though less reliable, suggest that explainability enhanced perceived predictability and confidence, especially for novices. 

Taken together, these findings highlight a central tension in designing XAI for no-code environments. Novices benefit from simple, visually intuitive explanations that build trust and transparency, while experts require depth, technical insights to diagnose errors and validate robustness. Balancing these divergent needs underscores the importance of adaptive explanation strategies that adjust depth and interpretability to user expertise and task context.

\section{Conclusions}
\label{conclusions}

We presented a user study ($N=20$, including both ML novices and experts) of an explainability module integrated into a no-code ML software. The module combines PDP, PFI, and KernelSHAP to provide global and local explanations within the workflow. Results indicate a consistently high task success rate ($\geq 80\%$) across all explainability tasks. Novices rated the explanations as useful, accurate, and trustworthy on the Explanation Satisfaction Scale (ESS, Cronbach’s $\alpha = 0.74$), whereas experts were more critical, particularly regarding sufficiency and completeness. Explanations also improved perceived predictability and confidence according to the Trust in Automation (TiA, $\alpha = 0.60$), with novices reporting higher levels of trust than experts. Overall, these findings highlight the importance of human-centered XAI in no-code platforms, where accessibility and trust-building for beginners must be balanced with diagnostic depth for expert users. Future work will extend this approach to explanations for large language models (LLMs), focusing on reasoning processes and interactive, context-aware interpretability to better support diverse user needs.

\section*{Acknowledgments}
This work was supported by ANID Basal Fund, National Center for Artificial Intelligence CENIA FB210017, Millennium Science Initiative Program Code ICN17\_002, Postdoctoral FONDECYT grant 3240001, FONDECYT regular grant 1231724, and ANID FONDEF grant ID25I10330.

\bibliographystyle{plainnat}
\bibliography{references}

@article{Alowais2023Revolutionizing,
title={Revolutionizing healthcare: the role of artificial intelligence in clinical practice},
author={Shuroug A. Alowais and Sahar S. Alghamdi and Nada Alsuhebany and Tariq Alqahtani and Abdulrahman I. Alshaya and Sumaya N Almohareb and Atheer Aldairem and Mohammed A. Alrashed and Khalid Bin saleh and H. Badreldin and Majed S Al Yami and Shmeylan A. Al Harbi and Abdulkareem M. Albekairy},
journal={BMC Medical Education},
year={2023},
volume={23},
doi={10.1186/s12909-023-04698-z}
}

@article{cao2022ai,
  title={Ai in finance: challenges, techniques, and opportunities},
  author={Cao, Longbing},
  journal={ACM Computing Surveys (CSUR)},
  volume={55},
  number={3},
  pages={1--38},
  year={2022},
  publisher={ACM New York, NY}
}

@article{magesh2025hallucination,
  title={Hallucination-Free? Assessing the Reliability of Leading AI Legal Research Tools},
  author={Magesh, Varun and Surani, Faiz and Dahl, Matthew and Suzgun, Mirac and Manning, Christopher D and Ho, Daniel E},
  journal={Journal of Empirical Legal Studies},
  volume={22},
  number={2},
  pages={216--242},
  year={2025},
  publisher={Wiley Online Library}
}

@article{rillig2024ai,
  title={Ai personal assistants and sustainability: Risks and opportunities},
  author={Rillig, Matthias C and Kasirzadeh, Atoosa},
  journal={Environmental Science Technology},
  volume={58},
  number={17},
  pages={7237--7239},
  year={2024},
  publisher={ACS Publications}
}

@article{Thalpage2023Unlocking,
title={Unlocking the Black Box: Explainable Artificial Intelligence (XAI) for Trust and Transparency in AI Systems},
author={Nipuna Thalpage},
journal={Journal of Digital Art Humanities},
year={2023},
doi={10.33847/2712-8148.4.1}
}

@article{Franzoni2023From,
title={From Black Box to Glass Box: Advancing Transparency in Artificial Intelligence Systems for Ethical and Trustworthy AI},
author={Valentina Franzoni},
year={2023},
pages={118-130},
doi={10.1007/978-3-031-37114-1},
journal={Computational Science and Its Applications}
}

@article{Hassija2023Interpreting,
title={Interpreting Black-Box Models: A Review on Explainable Artificial Intelligence},
author={Vikas Hassija and V. Chamola and Atmesh Mahapatra and Abhinandan Singal and Divyansh Goel and Kaizhu Huang and Simone Scardapane and Indro Spinelli and Mufti Mahmud and Amir Hussain},
journal={Cognitive Computation},
year={2023},
volume={16},
pages={45-74},
doi={10.1007/s12559-023-10179-8}
}

@article{Ananny2018Seeing,
title={Seeing without knowing: Limitations of the transparency ideal and its application to algorithmic accountability},
author={Mike Ananny and K. Crawford},
journal={New Media   Society},
year={2018},
volume={20},
pages={973 - 989},
doi={10.1177/1461444816676645}
}

@article{Burkart2020A,
title={A Survey on the Explainability of Supervised Machine Learning},
author={Nadia Burkart and Marco F. Huber},
journal={ArXiv},
year={2020},
volume={abs/2011.07876},
doi={10.1613/jair.1.12228}
}

@article{Wanner2022The,
title={The effect of transparency and trust on intelligent system acceptance: Evidence from a user-based study},
author={Jonas Wanner and L. Herm and K. Heinrich and Christian Janiesch},
journal={Electronic Markets},
year={2022},
volume={32},
pages={2079 - 2102},
doi={10.1007/s12525-022-00593-5}
}

@article{Naser2021An,
title={An engineer's guide to eXplainable Artificial Intelligence and Interpretable Machine Learning: Navigating causality, forced goodness, and the false perception of inference},
author={M. Z. Naser},
journal={Automation in Construction},
year={2021},
volume={129},
pages={103821},
doi={10.1016/J.AUTCON.2021.103821}
}

@article{Karim2022Explainable,
title={Explainable AI for Bioinformatics: Methods, Tools, and Applications},
author={Md. Rezaul Karim and Tanhim Islam and O. Beyan and C. Lange and Michael Cochez and Dietrich Rebholz-Schuhmann and Stefan Decker},
journal={Briefings in bioinformatics},
year={2022},
doi={10.48550/arXiv.2212.13261}
}

@article{Sundberg2023Democratizing,
title={Democratizing artificial intelligence: How no-code AI can leverage machine learning operations},
author={Leif Sundberg and J. Holmström},
journal={Business Horizons},
year={2023},
doi={10.1016/j.bushor.2023.04.003}
}

@article{Li2022How,
title={How Can No/Low Code Platforms Help End-Users Develop ML Applications? - A Systematic Review},
author={Luyun Li and Zhanwei Wu},
year={2022},
pages={338-356},
doi={10.1007/978-3-031-21707-4},
journal={HCI International}
}

@inproceedings{ribeiro2016should,
  title={" Why should i trust you?" Explaining the predictions of any classifier},
  author={Ribeiro, Marco Tulio and Singh, Sameer and Guestrin, Carlos},
  booktitle={Proceedings of the 22nd ACM SIGKDD international conference on knowledge discovery and data mining},
  pages={1135--1144},
  year={2016}
}

@article{lundberg2017unified,
  title={A unified approach to interpreting model predictions},
  author={Lundberg, Scott M and Lee, Su-In},
  journal={Advances in neural information processing systems},
  volume={30},
  year={2017}
}

@article{Bhatt2019Explainable,
title={Explainable machine learning in deployment},
author={Umang Bhatt and Alice Xiang and Shubham Sharma and Adrian Weller and Ankur Taly and Yunhan Jia and Joydeep Ghosh and R. Puri and J. Moura and P. Eckersley},
journal={Proceedings of the 2020 Conference on Fairness, Accountability, and Transparency},
year={2019},
doi={10.1145/3351095.3375624}
}

@article{Belle2020Principles,
title={Principles and Practice of Explainable Machine Learning},
author={Vaishak Belle and I. Papantonis},
journal={Frontiers in Big Data},
year={2020},
volume={4},
doi={10.3389/fdata.2021.688969}
}

@article{Herm2022Stop,
title={Stop ordering machine learning algorithms by their explainability! A user-centered investigation of performance and explainability},
author={L. Herm and K. Heinrich and Jonas Wanner and Christian Janiesch},
journal={Int. J. Inf. Manag.},
year={2022},
volume={69},
pages={102538},
doi={10.1016/j.ijinfomgt.2022.102538}
}

@article{Marcinkevics2023Interpretable,
title={Interpretable and explainable machine learning: A methods‐centric overview with concrete examples},
author={Ricards Marcinkevics and Julia E. Vogt},
journal={Wiley Interdisciplinary Reviews: Data Mining and Knowledge Discovery},
year={2023},
volume={13},
doi={10.1002/widm.1493}
}

@article{Kong2024Toward,
title={Toward Human-centered XAI in Practice: A survey},
author={Xiangwei Kong and Shujie Liu and Luhao Zhu},
journal={Mach. Intell. Res.},
year={2024},
volume={21},
pages={740-770},
doi={10.1007/s11633-022-1407-3}
}

@article{Al-Ansari2024User‐Centered,
title={User‐Centered Evaluation of Explainable Artificial Intelligence (XAI): A Systematic Literature Review},
author={Noor Al-Ansari and Dena Al-Thani and Reem S. Al-Mansoori},
journal={Human Behavior and Emerging Technologies},
year={2024},
doi={10.1155/2024/4628855}
}

@article{Liao2021Human-Centered,
title={Human-Centered Explainable AI (XAI): From Algorithms to User Experiences},
author={Q. Liao and Microsoft Research and Canada Kush and R. Varshney and Kush R. Varshney},
journal={ArXiv},
year={2021},
volume={abs/2110.10790},
doi={}
}

@article{Guo2024Explainability,
title={Explainability in JupyterLab and Beyond: Interactive XAI Systems for Integrated and Collaborative Workflows},
author={G. Guo and Dustin Arendt and A. Endert},
journal={ArXiv},
year={2024},
volume={abs/2404.02081},
doi={10.48550/arXiv.2404.02081}
}

@article{Hooker2019Unrestricted,title={Unrestricted permutation forces extrapolation: variable importance requires at least one more model, or there is no free variable importance},author={G. Hooker and L. Mentch and Siyu Zhou},journal={Statistics and Computing},year={2019},volume={31},doi={10.1007/s11222-021-10057-z}}

@article{Molnar2020Model-agnostic,
title={Model-agnostic Feature Importance and Effects with Dependent Features - A Conditional Subgroup Approach},
author={Christoph Molnar and Gunnar Konig and B. Bischl and Giuseppe Casalicchio},
journal={Data Min. Knowl. Discov.},
year={2020},
volume={38},
pages={2903-2941},
doi={10.1007/s10618-022-00901-9}
}

@article{Aas2019Explaining,
title={Explaining individual predictions when features are dependent: More accurate approximations to Shapley values},
author={K. Aas and Martin Jullum and Anders Løland},
journal={ArXiv},
year={2019},
volume={abs/1903.10464},
doi={10.1016/J.ARTINT.2021.103502}
}

@article{hoffman2018metrics,
  title={Metrics for explainable AI: Challenges and prospects},
  author={Hoffman, Robert R and Mueller, Shane T and Klein, Gary and Litman, Jordan},
  journal={arXiv preprint arXiv:1812.04608},
  year={2018}
}

@inproceedings{korber2018theoretical,
  title={Theoretical considerations and development of a questionnaire to measure trust in automation},
  author={K{\"o}rber, Moritz},
  booktitle={Congress of the International Ergonomics Association},
  pages={13--30},
  year={2018},
  organization={Springer}
}

@inproceedings{sundararajan2020many,
  title={The many Shapley values for model explanation},
  author={Sundararajan, Mukund and Najmi, Amir},
  booktitle={International conference on machine learning},
  pages={9269--9278},
  year={2020},
  organization={PMLR}
}

@article{Narayanan2018How,
title={How do Humans Understand Explanations from Machine Learning Systems? An Evaluation of the Human-Interpretability of Explanation},
author={Menaka Narayanan and Emily Chen and Jeffrey He and Been Kim and S. Gershman and F. Doshi-Velez},
journal={ArXiv},
year={2018},
volume={abs/1802.00682},
doi={}
}

@article{Sathyan2022Interpretable,
title={Interpretable AI for bio-medical applications},
author={Anoop Sathyan and Abraham Itzhak Weinberg and Kelly Cohen},
journal={Complex engineering systems (Alhambra, Calif.)},
year={2022},
volume={2},
doi={10.20517/ces.2022.41}
}

@article{2023Interpretable,
title={Interpretable Predictive Modeling of Tight Gas Well Productivity with SHAP and LIME Techniques},
author={Xianlin Ma and M. Hou and J. Zhan and Z. Liu},
journal={Energies},
year={2023},
doi={10.3390/en16093653}
}

@article{Chiesa-Estomba2023Explainable,
title={Explainable AI for Retinoblastoma Diagnosis: Interpreting Deep Learning Models with LIME and SHAP},
author={C. Chiesa-Estomba and M. Graña and Bader Aldughayﬁq and Farzeen Ashfaq and Noor Zaman Jhanjhi and M. Humayun},
journal={Diagnostics},
year={2023},
volume={13},
doi={10.3390/diagnostics13111932}
}

@article{Papenmeier2022It’s,
title={It’s Complicated: The Relationship between User Trust, Model Accuracy and Explanations in AI},
author={A. Papenmeier and Dagmar Kern and G. Englebienne and C. Seifert},
journal={ACM Transactions on Computer-Human Interaction (TOCHI)},
year={2022},
volume={29},
pages={1 - 33},
doi={10.1145/3495013}
}

@article{Pareek2024Effect,
title={Effect of Explanation Conceptualisations on Trust in AI-assisted Credibility Assessment},
author={Saumya Pareek and N. V. Berkel and Eduardo Velloso and Jorge Goncalves},
journal={Proceedings of the ACM on Human-Computer Interaction},
year={2024},
volume={8},
pages={1 - 31},
doi={10.1145/3686922}
}

@article{Westphal2023Decision,
title={Decision control and explanations in human-AI collaboration: Improving user perceptions and compliance},
author={Monika Westphal and Michael Vössing and G. Satzger and G. Yom-Tov and A. Rafaeli},
journal={Comput. Hum. Behav.},
year={2023},
volume={144},
pages={107714},
doi={10.1016/j.chb.2023.107714}
}

@incollection{Picard2023Human-Computer,
  title        = {Human-Computer Interaction and Explainability: Intersection and Terminology},
  author       = {Picard, Arthur and Mualla, Yazan and Gechter, Franck and Galland, Stéphane},
  booktitle    = {Explainable and Transparent AI and Multi-Agent Systems},
  editor       = {Calvaresi, Davide and Najjar, Amro and Winikoff, Michael and Fr\"{a}mling, Kary},
  publisher    = {Springer},
  year         = {2023},
  pages        = {214--236},
  doi          = {10.1007/978-3-031-44067-0}
}

@article{Reddy2024Human-Computer,
  title        = {Human-Computer Interaction Techniques for Explainable Artificial Intelligence Systems},
  author       = {Reddy, S. T. Anand},
  journal      = {Research Review: Machine Learning and Cloud Computing},
  volume       = {3},
  number       = {1},
  pages        = {1--7},
  year         = {2024},
  doi          = {10.46610/rtaia.2024.v03i01.001}
}

@article{Mohseni2018A,
title={A Multidisciplinary Survey and Framework for Design and Evaluation of Explainable AI Systems},
author={Sina Mohseni and Niloofar Zarei and E. Ragan},
journal={ACM Trans. Interact. Intell. Syst.},
year={2018},
volume={11},
pages={24:1-24:45},
doi={10.1145/3387166}
}

@article{Rong2022Towards,
title={Towards Human-Centered Explainable AI: A Survey of User Studies for Model Explanations},
author={Yao Rong and Tobias Leemann and Thai-trang Nguyen and Lisa Fiedler and Peizhu Qian and Vaibhav Unhelkar and Tina Seidel and Gjergji Kasneci and Enkelejda Kasneci},
journal={IEEE Transactions on Pattern Analysis and Machine Intelligence},
year={2022},
volume={46},
pages={2104-2122},
doi={10.1109/TPAMI.2023.3331846}
}

@article{Tian2024Automated,
title={Automated Machine Learning: A Survey of Tools and Techniques},
author={Junchi Tian and Chang Che},
journal={Journal of Industrial Engineering and Applied Science},
year={2024},
doi={10.70393/6a69656173.323336}
}

@article{Nguyen2024XEdgeAI,
title={XEdgeAI: A Human-centered Industrial Inspection Framework with Data-centric Explainable Edge AI Approach},
author={Truong Thanh Hung Nguyen and Phuc Truong Loc Nguyen and Hung Cao},
journal={Inf. Fusion},
year={2024},
volume={116},
pages={102782},
doi={10.48550/arXiv.2407.11771}
}

@article{Cambria2023A,
title={A survey on XAI and natural language explanations},
author={E. Cambria and Lorenzo Malandri and Fabio Mercorio and Mario Mezzanzanica and Navid Nobani},
journal={Inf. Process. Manag.},
year={2023},
volume={60},
pages={103111},
doi={10.1016/j.ipm.2022.103111}
}

@article{Bodria2021Benchmarking,
title={Benchmarking and survey of explanation methods for black box models},
author={F. Bodria and F. Giannotti and Riccardo Guidotti and Francesca Naretto and D. Pedreschi and S. Rinzivillo},
journal={Data Mining and Knowledge Discovery},
year={2021},
volume={37},
pages={1719-1778},
doi={10.1007/s10618-023-00933-9}
}

@article{Waa2021Evaluating,
title={Evaluating XAI: A comparison of rule-based and example-based explanations},
author={J. V. D. Waa and Elisabeth Nieuwburg and A. Cremers and Mark Antonius Neerincx},
journal={Artif. Intell.},
year={2021},
volume={291},
pages={103404},
doi={10.1016/j.artint.2020.103404}
}

@article{graells2025tsundoku,
  title={Tsundoku: A Python toolkit for social network analysis},
  author={Graells-Garrido, Eduardo and Garc{\'\i}a, Nicol{\'a}s and Carvallo, Andr{\'e}s},
  journal={SoftwareX},
  volume={29},
  pages={102008},
  year={2025},
  publisher={Elsevier}
}

@inproceedings{parra2019analyzing,
  title={Analyzing the design space for visualizing neural attention in text classification},
  author={Parra, D and Valdivieso, H and Carvallo, A and Rada, G and Verbert, K and Schreck, T},
  booktitle={Proc. ieee vis workshop on vis x ai: 2nd workshop on visualization for ai explainability (visxai)},
  year={2019}
}

@inproceedings{carvallo2025user_final,
  title={User Perception of Attention Visualizations: Effects on Interpretability Across Evidence-Based Medical Documents},
  author={Carvallo, Andr{\'e}s and Parra, Denis and Brusilovsky, Peter and Valdivieso, Hernan and Rada, Gabriel and Donoso, Ivania and Araujo, Vladimir},
  booktitle={International Workshop on Human-AI Collaboration},
  pages={71--80},
  year={2025},
  organization={Springer}
}

\section*{Appendix}

\appendix

\section{User test tasks}
\label{annex:tasks}

Table~\ref{tab:tasks-annex} summarizes the eleven tasks designed for the user study. While all participants completed the full workflow, only Tasks 7–11 were directly related to the explainability module.

\begin{table}[ht]
\centering
\caption{User test tasks and success criteria.}
\label{tab:tasks-annex}
\scriptsize
\begin{tabularx}{\textwidth}{c X c X}
\toprule
\textbf{ID} & \textbf{Task} & \textbf{Scenario} & \textbf{Success Criterion} \\
\midrule
T1 & Load the \textit{Twitter Sentiment Analysis} dataset & A and B & Dataset appears in the interface table \\
T2 & Train a text classification model on the \textit{Twitter Sentiment Analysis} dataset & A and B & Experiment runs successfully and appears in the interface \\
T3 & Check results of the text classification model & A and B & User opens the tab showing the results \\
T4 & Load the \textit{customer churn} dataset and explore it & A and B & User opens the modal showing an example of the dataset \\
T5 & Create an experiment with two KNN models & A and B & Experiment runs successfully and appears in the interface \\
T6 & Check results of the KNN models & A and B & User opens the tab showing the results \\
T7 & Select a KNN model and open its explainability dashboard & A and B & User accesses the explainability interface for the chosen model \\
T8 & Implement PFI and analyze results & A & Explanation is displayed and user interprets it aloud \\
T9 & Implement PDP and analyze results & A & Explanation is displayed and user interprets it aloud \\
T10 & Implement KernelSHAP and analyze results & A & Explanation is displayed and user interprets it aloud \\
T11 & Implement PFI, PDP, and KernelSHAP and analyze results jointly & B & Explanations are displayed and user interprets them aloud \\
\bottomrule
\end{tabularx}
\end{table}

\section{Explanation Satisfaction Scale (ESS)}
\label{annex:ess}

The ESS questionnaire was administered to all participants. It contains eight statements that assess user satisfaction with model explanations. Each statement is rated on a 5-point Likert scale: 

\begin{center}
\textit{Strongly disagree, Disagree, Neither agree nor disagree, Agree, Strongly agree}
\end{center}

\noindent The full questionnaire is reproduced below:

\begin{enumerate}
    \item Based on the explanations, I understand how the trained model works.  
    \item The explanations of how the trained model works are satisfactory to me.  
    \item The explanations of how the trained model works contain sufficient details.  
    \item The explanations of how the trained model works appear complete to me.  
    \item The explanations of how the trained model works tell me how to use it.  
    \item The explanations of how the trained model works are useful for the intended goals.  
    \item The explanations of how the trained model works show me how accurate it is.  
    \item The explanations of the trained model allow me to judge when I should and should not trust it.  
\end{enumerate}

\section{Trust in Automation (TiA)}
\label{annex:tia}

The TiA questionnaire was adapted from Korber et. al\citep{korber2018theoretical}. It included four subscales: 
\textit{familiarity}, \textit{predictability}, \textit{propensity to trust}, and \textit{confidence in automation}. Participants rated each item on a 5-point Likert scale (1 = Strongly Disagree, 5 = Strongly Agree).  

The evaluated items are as follows: 
\begin{enumerate}
    \item I am familiar with systems similar to this one. 
    \item I can easily predict the system’s behavior based on my interactions. 
    \item I usually tend to trust automated systems. 
    \item I feel confident in relying on the model’s outputs after receiving explanations. 
    \item The system behaves in a way that I consider consistent with my expectations. 
    \item I would recommend using this system to others in similar tasks. 
    \item I am inclined to delegate decisions to this system when explanations are provided. 
    \item The system demonstrates competence in the tasks it was designed for. 
    \item Explanations make it easier to decide when to trust or not trust the system. 
    \item Overall, I feel comfortable trusting the automation in this software. 
\end{enumerate}

\section{Structured Interview on Explainability Module}
\label{annex:structured_interview}

The structured interviews conducted in the study follow a predefined sequence. For the purposes of this annex, we present only the explainability-related tasks (T7 onwards):

\begin{enumerate}
    \item To familiarize participants with the explainability module and techniques, a demonstration video is shown. This video illustrates how to configure the explanation tools in the interface and how to interpret the results of each of the three implemented methods (PDP, PFI, and SHAP).
    
    \item Depending on the assigned scenario, participants are asked to complete either tasks T7 to T10, or tasks T7 and T11. These tasks specifically involve interacting with the explanation methods and applying them to different model outputs.
    
    \item Based on their interaction with the explainability interface, participants are asked to complete the ESS questionnaire and the following TiA subscales: 
    \begin{itemize}
        \item System comprehension/predictability
        \item Trust in automation
    \end{itemize}
    
    \item Considering the complete experience with DashAI, participants are asked to complete the SUS questionnaire and the following TiA subscales: 
    \begin{itemize}
        \item Familiarity with similar systems
        \item Propensity to trust an autonomous system
    \end{itemize}
    
    \item Finally, participants answer a set of open-ended questions to provide qualitative feedback on their experience with the explainability module.
\end{enumerate}

\end{document}